# Facial Expression Recognition using Visual Saliency and Deep Learning

Viraj Mavani
L. D. College of Engineering
Ahmedabad
mavani.viraj.604@ldce.ac.in

Shanmuganathan Raman
Indian Institute of Technology
Gandhinagar
shanmuga@iitgn.ac.in

Krishna P Miyapuram
Indian Institute of Technology
Gandhinagar
kprasad@iitgn.ac.in

## Abstract

*We have developed a convolutional neural network for the purpose of recognizing facial expressions in human beings. We have fine-tuned the existing convolutional neural network model trained on the visual recognition dataset used in the ILSVRC2012 to two widely used facial expression datasets - CFEE and RaFD, which when trained and tested independently yielded test accuracies of 74.79% and 95.71%, respectively. Generalization of results was evident by training on one dataset and testing on the other. Further, the image product of the cropped faces and their visual saliency maps were computed using Deep Multi-Layer Network for saliency prediction and were fed to the facial expression recognition CNN. In the most generalized experiment, we observed the top-1 accuracy in the test set to be 65.39%. General confusion trends between different facial expressions as exhibited by humans were also observed.*

## 1. Introduction

Human facial expressions play a vital role in regular interaction between human beings. Facial expressions are key to recognize human emotions. The goal of having state-of-the-art machine vision systems that can match humans has been pursued for a very long time now. For such applications, facial expression recognition becomes crucial.

Deep learning algorithms have been proven to be excellent for computer vision tasks like object detection and classification. Convolutional Neural Networks [12, 13] were developed to ease the process of feature selection and give better results than already existing machine learning methods. CNN architectures have developed to such a state that they even out-perform humans in various image classification tasks. The Convolutional Neural Network called the AlexNet [10] has 5 convolution layers and 60 million parameters. Deep Learning networks can be applied to classify facial expressions as well. For limiting extraneous data, we first need to crop the face regions from images. Grayscale facial images were given as training data to AlexNet and the results were observed.

Human facial expressions have been recognized time and again by various different methods mostly using the Facial Action Coding System (FACS) [6] as a guide. Predicting expressions based on FACS requires to detect action units from the image first using methods like facial feature point tracking, dense flow tracking or using gray-value changes. The data extracted this way is then forward propagated through the classifier for facial expression recognition. These approaches are computationally expensive as well as require many different blocks before the actual prediction is made. We propose a new method which implements a deep learning algorithm to recognize facial expressions using the image product of cropped faces and their detected visual saliency.

Visual Saliency is intensity map wherein higher intensities show those areas of an image which attract maximum attention and decreasing attention results in lower intensities. The visual saliency has been studied for a long time by scientists working on cognitive science and various methods have been developed to calculate the visual saliency of an image. These include gaze trackers with machine learning applied to the acquired data. These methods prove to be good while computing saliency maps. Newer approaches like Deep learning algorithms have shown better results in visual saliency prediction [8].

The product of the saliency map and the cropped face signifies the image as a normal human being perceives it and only those parts of the image is taken into consideration which demand visual attention. For this we have used different pre-processing techniques and a Deep Multi-Layer Network for saliency prediction [2]. After cropping out faces and computing the saliency maps which are basically intensity maps, we multiply them both. The result of this operation is then passed on to the deep convolutional neural network AlexNet [10] that has been pre-trained on the data-set for object classification from the competition ILSVRC 2012 [21]. We fine-tuned this network for facial expression recognition using two different datasets. We used the Radboud University's Faces Database (RaFD) [11] and the Compound Facial

Expressions of Emotion Dataset [5]. We trained the network on CFEE dataset and tested the trained model on RaFD.

## 2. Related Work

Much work has been done for facial expression recognition using the Facial Action Coding System [6]. This has been clearly showcased in works like [1, 19, 25]. FACS classifies emotions on the basis of different Action Units (AU) on the human face. Several action units form a single expression. These approaches have showcased great results as well. Other approaches like Bayesian Networks, Artificial Neural Networks and Hidden Markov Model (HMM) have also been used for facial expression recognition [14, 17]. We have instead used a Convolutional Neural Network to classify these images and evaluated the results. CNN does not use the key features described by the FACS.

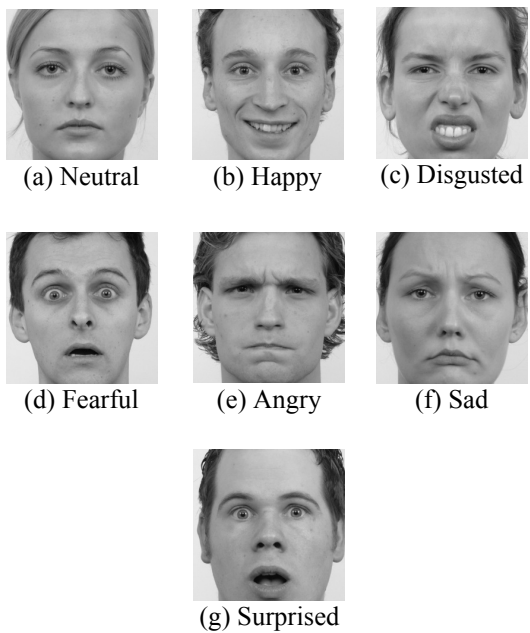

(a) Neutral    (b) Happy    (c) Disgusted

(d) Fearful    (e) Angry    (f) Sad

(g) Surprised

Figure 1: Sample of images from the datasets

To understand the deep learning approach, we need to first look into the general approach used for facial expression recognition. Firstly, image registration is done. Registration requires the faces in the image to be localized using face detection algorithms [26] and then matched to a template image. After that, feature extraction algorithms such as the Histogram of Oriented Gradients (HOG) [3], Local Binary Patterns (LBP) [23], Gabor filters [15] are applied to these registered images. The algorithm produces a feature vector which is then fed to the classifier for classification.

Deep Convolutional Neural Networks on the other hand do not require us to define the feature extraction algorithms to be used. While training, the network itself learns the weights and biases as well as the kernels to be convoluted with the image for feature extraction. Such approaches have been used in the past in [4, 16, 27]. These approaches vary greatly in terms of the CNN architecture used and also on methodology.

## 3. The Data

We have used the CFEE and the RaFD datasets. 7 basic emotions from both the datasets were extracted and categorized in to different folders. These seven emotions are Angry, Fearful, Disgusted, Surprised, Happy, Sad and Neutral. The faces from all the images of both the datasets were cropped out using the Viola Jones Detector for faces [26]. These images were then converted to 256x256 in size and the color channel was changed to grayscale.

The CFEE dataset has a total number of 1610 images for 230 subjects. The RaFD dataset has a total of 1407 images for 67 subjects with each subject having 3 different gaze directions i.e. front, left and right. [11] is a well posed facial expressions dataset with the eyes of all the subjects aligned together. This makes the data extremely coherent and prediction becomes easier. This is not the case in real time. We did multiple experiments with different data splits.

## 4. Experiments

We have done various experiments to compare the performance of the AlexNet [10] on the datasets. We fine-tuned the network which was pre-trained on the ILSVRC2012 competition dataset [21] for image classification to the datasets. We also tuned the hyper-parameters like learning rate, decay rate, regularization and dropout [24] during experimentation. A softmax [9] layer of 7 outputs was applied at the end of the model.

We tested with different data splits and in the end tested it by computing the visual saliency maps of each of the images and taking its product image as training data. We also used Domain Adaptation techniques [20] to improve generalization of the models.

Such experimentation helped us in getting insight on the role of human gaze in simple visual tasks like classification.

|  | Angry | Disgusted | Fearful | Happy | Neutral | Sad | Surprised | Per-class Accuracy |
|---|---|---|---|---|---|---|---|---|
| **Angry** | 114 | 0 | 0 | 0 | 3 | 84 | 0 | 56.72% |
| **Disgusted** | 8 | 166 | 0 | 2 | 20 | 1 | 4 | 82.59% |
| **Fearful** | 0 | 0 | 95 | 0 | 58 | 16 | 32 | 47.26% |
| **Happy** | 2 | 0 | 2 | 187 | 7 | 3 | 0 | 93.03% |
| **Neutral** | 5 | 0 | 0 | 0 | 135 | 59 | 2 | 67.16% |
| **Sad** | 3 | 0 | 0 | 0 | 10 | 188 | 0 | 93.53% |
| **Surprised** | 0 | 0 | 0 | 0 | 0 | 0 | 201 | 100.0% |

Table 1: The confusion matrix of AlexNet trained on CFEE and tested on RaFD

### 4.1. Training and Testing on the CFEE

We used the CFEE Dataset and extracted the seven facial expressions for training and testing both. The dataset contains 1610 images in total. We kept 1127 images for training, 245 images in the cross validation set and 238 images in the test set. This makes it an approximate split of 70%, 15% and 15%. There exists no overlap between the training and test images. We achieved a test accuracy of 74.79%. This is because the test set consist images of the same subjects as that of the training set but different facial expressions. This result cannot be considered to be much generalized.

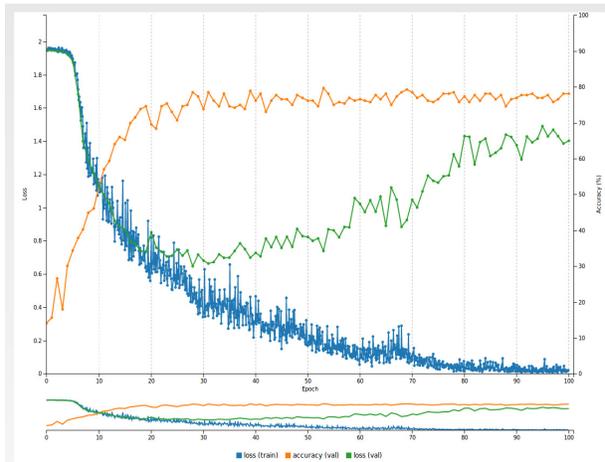

Figure 2: Training Curve for 4.1
Blue – Training Loss
Green – Validation Loss
Orange – Validation Accuracy

### 4.2. Training and Testing on RaFD

We also used the RaFD to train the AlexNet. It contains 1407 images in total of 67 subjects with 3 images each. The 3 images of each subject have different gaze directions. The training was done on 987 images, the validation set had 210 images and the test set consisted of 210 images. There was no overlap between any of these sets. It is an approximate split of 70% training, 15% validation and 15% test. Dropout was used to improve generalization. On testing, the accuracy was found to be 95.71%. This is an exceptionally good result quantitatively but still lacks in generalization.

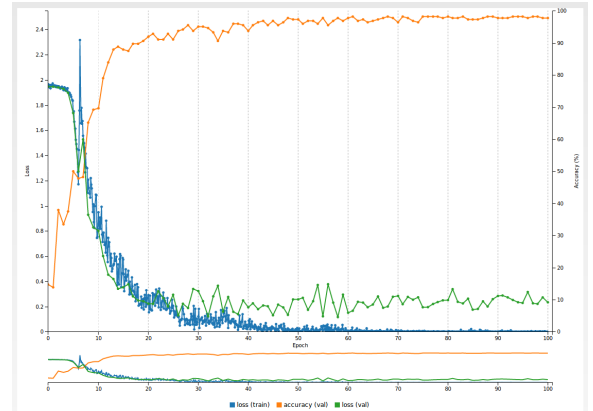

Figure 3: Training Curve for 4.2
Blue – Training Loss
Green – Validation Loss
Orange – Validation Accuracy

### 4.3. Training on CFEE and Testing on RaFD

For better generalization, we trained the network on the full CFEE Dataset i.e. 1610 images. As using different datasets for training and testing improves generalization because of the difference in the way of acquisition, the environment, etc., this model would give us proper results in terms of generalizability. This indeed provided better

|  | Angry | Disgusted | Fearful | Happy | Neutral | Sad | Surprised | Per-class Accuracy |
|---|---|---|---|---|---|---|---|---|
| **Angry** | 93 | 78 | 0 | 0 | 8 | 22 | 0 | 46.27% |
| **Disgusted** | 16 | 181 | 0 | 2 | 2 | 0 | 0 | 90.05% |
| **Fearful** | 8 | 17 | 131 | 2 | 12 | 11 | 20 | 65.17% |
| **Happy** | 2 | 29 | 2 | 160 | 0 | 3 | 5 | 79.6% |
| **Neutral** | 17 | 41 | 4 | 1 | 103 | 31 | 4 | 51.24% |
| **Sad** | 22 | 50 | 2 | 1 | 29 | 93 | 4 | 46.27% |
| **Surprised** | 0 | 2 | 38 | 0 | 0 | 2 | 159 | 79.1% |

Table 2: The confusion matrix of AlexNet trained on CFEE Image Saliency Products and tested on RaFD Image Saliency Products

generalization. The model was tested on the RaFD and a test accuracy of 77.19% was achieved.

The model was optimized using a stochastic gradient descent algorithm with a base learning rate of 0.01. We also applied a linear learning rate decay spread through the 100 epochs for which the training was done. The training of the caffemodel was done using the NVIDIA DIGITS Deep Learning framework [18]. This model provides satisfactory results when it is to be put to use for real time applications.

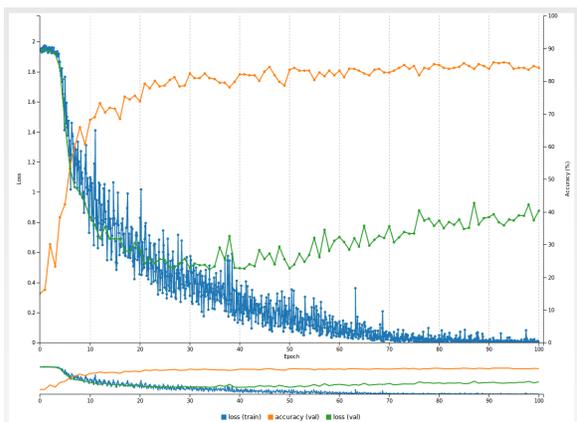

Figure 4: Training Curve for 4.3
Blue – Training Loss
Green – Validation Loss
Orange – Validation Accuracy

### 4.4. Introducing Visual Saliency

#### 4.4.1 Computing the Visual Saliency Maps

The visual saliency maps of the images of both the dataset were computed using the pre-trained CNN given in Deep Multi-Level Convolutional Neural Network [2]. ML-Net describes a new CNN architecture for saliency prediction. It contains a generic feature extraction CNN, a feature encoding network and a prior learning network.

Ml-Net outperforms all the other models on the SALICON Dataset [7] and also performs reasonably well on the MIT Saliency Benchmark [8]. These maps were generated using a script which iterated over the entire dataset feed-forwarding each image trough the network.

#### 4.4.2 Computing the product image and visual saliency map respectively

A python script was used to compute the image product with pixel scaling of the visual saliency maps and the images respectively. The result of the process is shown in figure 5.

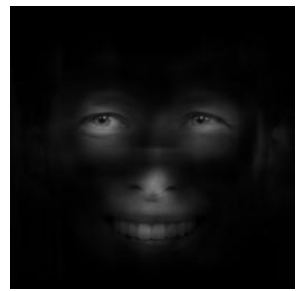

Figure 5: Scaled Image Product

#### 4.4.3 Training on CFEE Image Products and Testing on RaFD Image Products

We trained the network using the image products of saliency maps and respective images of the CFEE dataset and tested it on its RaFD counterpart. The results were a bit intriguing. The test accuracy was found to be 65.39%. This model was also optimized using a stochastic gradient descent algorithm with a base learning rate of 0.01 accompanied with linear learning rate decay. The model was trained for 100 epochs. The training curve is given in figure 6. The confusion matrix can be seen in Table 2.

## 5. Discussion and Conclusion

We have demonstrated a CNN for facial expression recognition with generalization abilities. We tested the contribution of potential facial regions of interest in human vision using visual saliency of images in our facial expressions datasets.

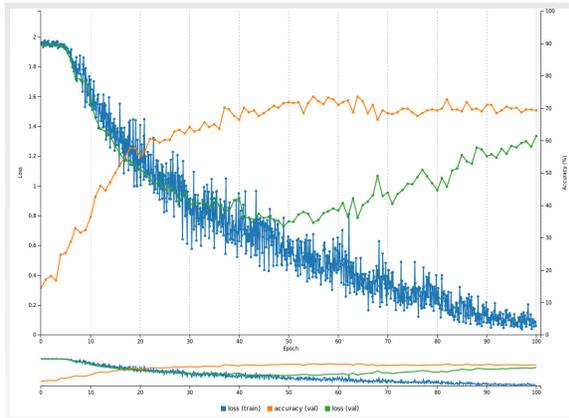

Figure 6: Training Curve for 4.3
Blue – Training Loss
Green – Validation Loss
Orange – Validation Accuracy

The confusion between different facial expressions was minimal with high recognition accuracies for four emotions – disgust, happy, sad and surprise [Table 1, 2]. The general human tendency of angry being confused as sad was observed [Table 1] as given in [22]. Fearful was confused with neutral, whereas neutral was confused with sad. When saliency maps were used, we observed a change in the confusion matrix of emotion recognition accuracies. Angry, neutral and sad emotions were now more confused with disgust, whereas surprised was more confused as fearful [Table 2]. These results suggested that the generalization of deep learning network with visual saliency 65.39% was much higher than chance level of 1/7. Yet, the structure of confusion matrix was much different when compared to the deep learning network that considered complete images. We conclude with the key contributions of the paper as two-fold. (i), we have presented generalization of deep learning network for facial emotion recognition across two datasets. (ii), we introduce here the concept of visual saliency of images as input and observe the behavior of the deep learning network to be varied. This opens up an exciting discussion on further integration of human emotion recognition (exemplified using visual saliency in this paper) and those of deep convolutional neural networks for facial expression recognition.

## 6. Future Work

This work can be carried forward by studying the human performance for each facial expression on both the datasets. Comparison between the CNN model and the human performance should then be evaluated using different metrics. The results would enable us to see how well the model performs in comparison general human abilities.

## 7. Acknowledgments

We are grateful to the anonymous reviewers of this paper for their wise comments because of which we were able to improve the quality of the paper and to Prof. Usha Neelakantan, Head of Department, L. D. College of Engineering for her support.